\documentclass{article}[12pt]

\usepackage[utf8]{inputenc}

\usepackage[margin=20mm]{geometry}
\usepackage{setspace}
\usepackage[english]{babel}
\usepackage{natbib}
\usepackage{amsmath,amssymb,amsfonts}
\usepackage{algorithmic}
\usepackage{graphicx}
\usepackage{textcomp}
\usepackage{xcolor}
\usepackage{multirow}
\usepackage{pdflscape}

\usepackage{acronym}
\usepackage{array,multirow}
\usepackage{float}
\usepackage{subfig}

\acrodef{CNN}{Convolutional Neural Network}

\begin{document}
	
\title{Automatic image-based identification and biomass estimation of invertebrates}
	
\author{Johanna Ärje$^{a, b, c}$, Claus Melvad$^d$, Mads Rosenh\o j Jeppesen$^d$, Sigurd Agerskov Madsen$^d$,\\ Jenni Raitoharju$^e$, Maria Strandg\aa rd Rasmussen$^b$,\\ Alexandros Iosifidis$^f$, Ville Tirronen$^g$, Kristian Meissner$^e$, Moncef Gabbouj$^a$, Toke Thomas H\o ye$^b$\\\\
$^a$ Unit of Computing Sciences, Tampere University, Finland.\\
$^b$ Department of Bioscience, Aarhus University, Denmark\\
$^c$ Department of Mathematics and Statistics, University of Jyvaskyla, Finland\\
$^d$ Aarhus School of Engineering, Aarhus University, Denmark\\
$^e$ Finnish Environment Institute, Jyväskylä, Finland\\
$^f$ Department of Engineering, Aarhus University, Denmark\\
johanna.arje@tuni.fi}
	
\maketitle

\newpage
\begin{abstract}
	Understanding how biological communities respond to environmental changes is a key challenge in ecology and ecosystem management. The apparent decline of insect populations necessitates more biomonitoring but the time-consuming sorting and identification of taxa pose strong limitations on how many insect samples can be processed. In turn, this affects the scale of efforts to map invertebrate diversity altogether. Given recent advances in computer vision, we propose to replace the standard manual approach of human expert-based sorting and identification with an automatic image-based technology. We describe a robot-enabled image-based identification machine, which can automate the process of invertebrate identification, biomass estimation and sample sorting. We use the imaging device to generate a comprehensive image database of terrestrial arthropod species. We use this database to test the classification accuracy i.e. how well the species identity of a specimen can be predicted from images taken by the machine. We also test sensitivity of the classification accuracy to the camera settings (aperture and exposure time) in order to move forward with the best possible image quality. We use state-of-the-art Resnet-50 and InceptionV3 \acp{CNN} for the classification task. The results for the initial dataset are very promising ($\overline{ACC}=0.980$). The system is general and can easily be used for other groups of invertebrates as well. As such, our results pave the way for generating more data on spatial and temporal variation in invertebrate abundance, diversity and biomass.
\end{abstract}
	
\section*{keywords}
Beetles, Biodiversity, Classification, CNN, Deep Learning, Machine Learning, Spiders

\section{Introduction}

The uncertanties around the state of global insect populations is largely due to data gaps and more efficient methods for quantifying abundance and identifying invertebrates are urgently needed \citep{seibold2019, wagner2020}. Commonly used passive traps, such as the Malaise traps produce samples, which are time consuming to process. For this reason, samples are sometimes only weighed - as was the case in the study, which triggered the global attention around insect declines \citep{hallmann2017}. In other studies, specimens are lumped into larger taxonomic groups \citep{timms2012,hoye2008,rich2013}, or only specific taxa are identified \citep{loboda2017,hansen2016}. On the other hand, such traps help standardise efforts across sampling events and are often preferred in long-term monitoring. Hence, the time and expertise needed to process (sort, identify, count and potentially weigh) samples of insects and other invertebrates from passive traps remains a key bottleneck in entomological research. In light of the apparent global decline of many invertebrate taxa, and the Linnean shortfall \citep[that only a small fraction of all species on Earth are described;][]{hortal2015}, more efficient ways of processing invertebrate samples are in high demand. Such methods should ideally 1) not destroy specimens, which could be new to the study area or even new to science, 2) count the abundance of individual species, and 3) estimate the biomass of such samples. 

Reliable identification of species is pivotal but due to its inherent slowness and high costs, traditional manual identification has caused bottlenecks in the bioassessment process. As the demand for biological monitoring grows, and the number of taxonomic experts declines \citep{gaston2004}, there is a need for alternatives to the manual processing and identification of monitoring samples \citep{borja2013, nygard2016}. While genetic approaches are gaining popularity and becoming standard tools in diversity assessments \citep{raupach2010, keskin2014, dunker2016, aylagas2016, kermarrec2014, elbrecht2017, zimmermann2015} they are still expensive and are not yet suitable to produce reliable abundance data or estimates of biomass. In stead, machine learning methods could be used to replace or semi-automate the task of manual species identification.

Several computer-based identification systems for biological monitoring have been proposed and tested in the last two decades. While \citet{potamis2014} has classified birds based on sound and \citet{qian2011} have used acoustic signals to identify bark beetles, most computer-based identification systems use morphological features and image data for species prediction. \citet{schroder1995, weeks1997, liu2008, lequing2012, perre2016} and \citet{feng2016} have classified bees, butterflies, fruit flies and wasps based on features calculated from their wings. In aquatic research, automatic or semi-automatic systems have been developed to identify algae \citep[e.g.][]{santhi2013}, zooplankton \citep[e.g.][]{dai2016, bochinski2018} and benthic macroinvertebrates \citep[e.g.][]{raitoharju2019b, arje2020}. In recent years, iNaturalist, a citizen-science application and community for recoding and sharing nature observations, has accumulated a notable database of taxa images for training state-of-the-art \acp{CNN} \citep{vanhorn2018}. However, such field photos will not provide the same accuracy as can be achieved in the lab under controlled light conditions.

Classification based on single 2D images can suffer from variations of the viewing angle and certain morphological traits being omitted. To overcome those limitations \citet{zhang2010} have proposed a method for structuring 3D insect models from 2D images. \citet{raitoharju2018} have presented an imaging system producing multiple images from two different angles for benthic macroinvertebrates. Using this latter imaging device and deep \acp{CNN}, \citet{arje2020} have achieved classification accuracy within the range of human taxonomic experts. 

Our aim for this work was 1) to make a reproducible imaging system, 2) to test the importance of different camera settings 3) to evaluate overall classification accuracy, and 4) to test the possibility of deriving biomass straight from geometrical features in images. To obtain these objectives, we rebuilt the imaging system presented in \citet{raitoharju2018} using industry components to make it completely reproducible. It has been made light proof to prevent false light from affecting the images. We also developed a flushing mechanism to pass specimens through the imaging device. This is a critical improvement for automation as explained below.
For classification, we used Resnet-50 \citep{he2016} and InceptionV3 \citep{szegedy2016} \acp{CNN}. We tested different camera settings (exposure time and aperture) to find the optimal settings for species identification, and we explored the necessary number of images per specimen to achieve high classification accuracy. Finally, for a subset of species, we tested if the area of a specimen derived directly from images taken by the device could serve as a proxy for biomass (dry weight) of the specimen.


\section{Materials and Methods}

\begin{figure}[htbp]
	\centering
	\subfloat[]{
		\includegraphics[width=8cm]{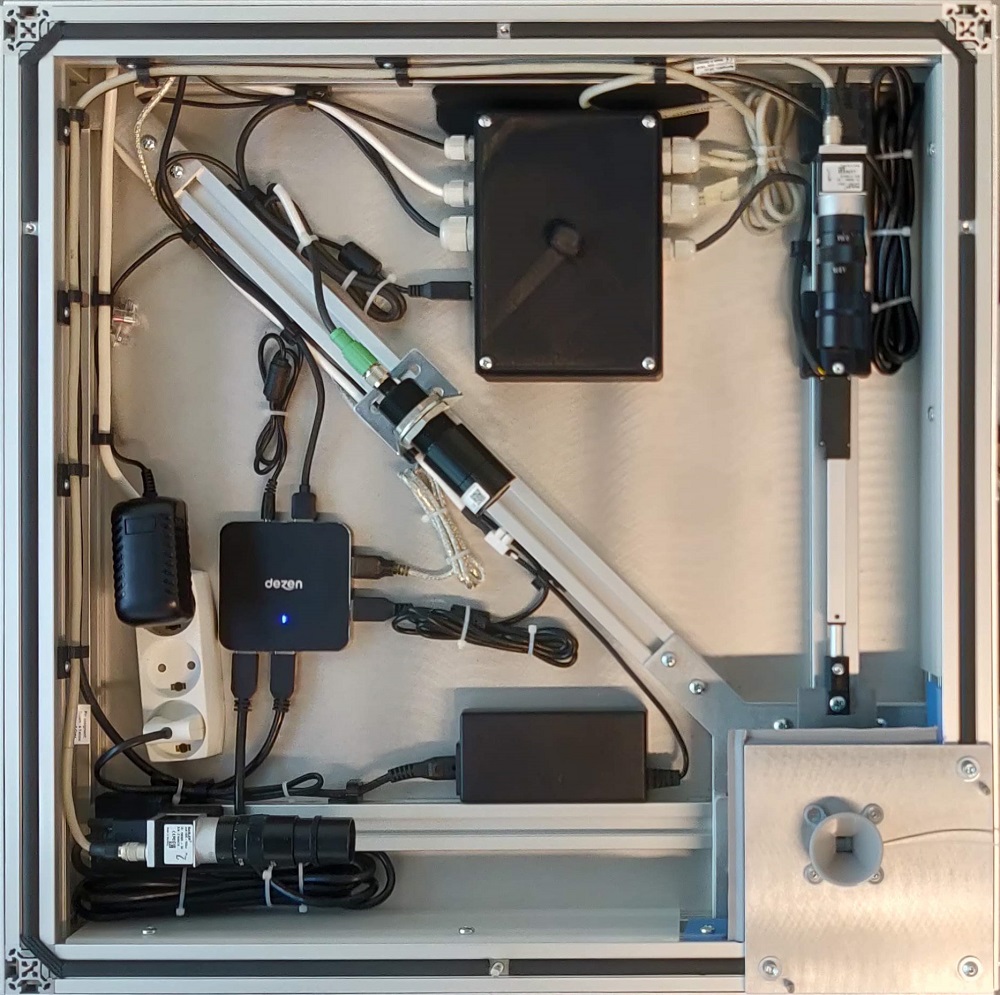}
		\label{inside}
	}
	\subfloat[]{
		\includegraphics[width=8cm]{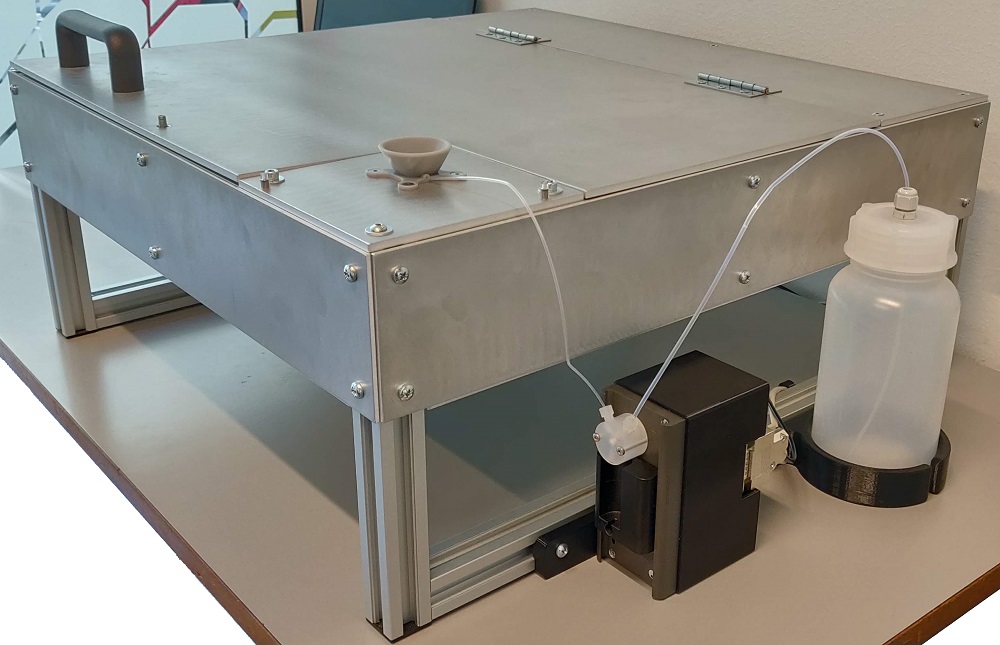}
		\label{outside}
	}\\
	\caption{BIODISCOVER machine for imaging invertebrates.}
	\label{BIODISCOVER}
\end{figure}

To facilitate the automation of specimen identification, biomass estimation and sorting of invertebrate specimens, we improved the prototype imaging system developed for automatic identication of benthic macroinvertebrates \citep{raitoharju2018}. We named the new device the BIODISCOVER machine, as an acronym for BIOlogical specimens Described, Identified, Sorted, Counted, and Observed using Vision-Enabled Robotics. The system comprises an aluminium box with two Basler ACA1920-155UC cameras and LD75 lenses with xo.15 to xo.35 magnification and five aperture settings (maximum aperture ratio of 1:3.8). The cameras are placed at a 90 degree angle to each other at two corners of the case and in the other corners there is a high power LED light (ODSX30-WHI Prox Light, which enables a maximum frame rate of 100 per second with exposure 1000) and a rectangular cuvette made of optical glass and filled with ethanol. The inside of the case is depicted in Fig. \ref{inside}. The case is rubber-sealed and has a lid to minimize light, shadows and other disturbances. The lid has an opening for the cuvette with a funnel for dropping specimens into the liquid. Fig. \ref{outside} shows the new refill system, which pumps ethanol into the cuvette.

The multiview imaging component is connected to a computer with an integrated software, which controls all parts of the machine. The program uses calibration images to detect objects differing from the background and triggers the light and cameras to take images as the specimen sinks in the ethanol until it disappears from the assigned view point of the cameras. The program detects the specimen and crops the images to be 496 pixels wide (defined by the width of the cuvette) and 496 pixels high while keeping the specimen at the center of the image with regards to the height. If a specimen exceeds the height of 496 pixels, the resulting images will be higher. The images are stored onto the computer as PNG files.

\begin{figure}[htbp]
	\centering
	\includegraphics[width=8cm]{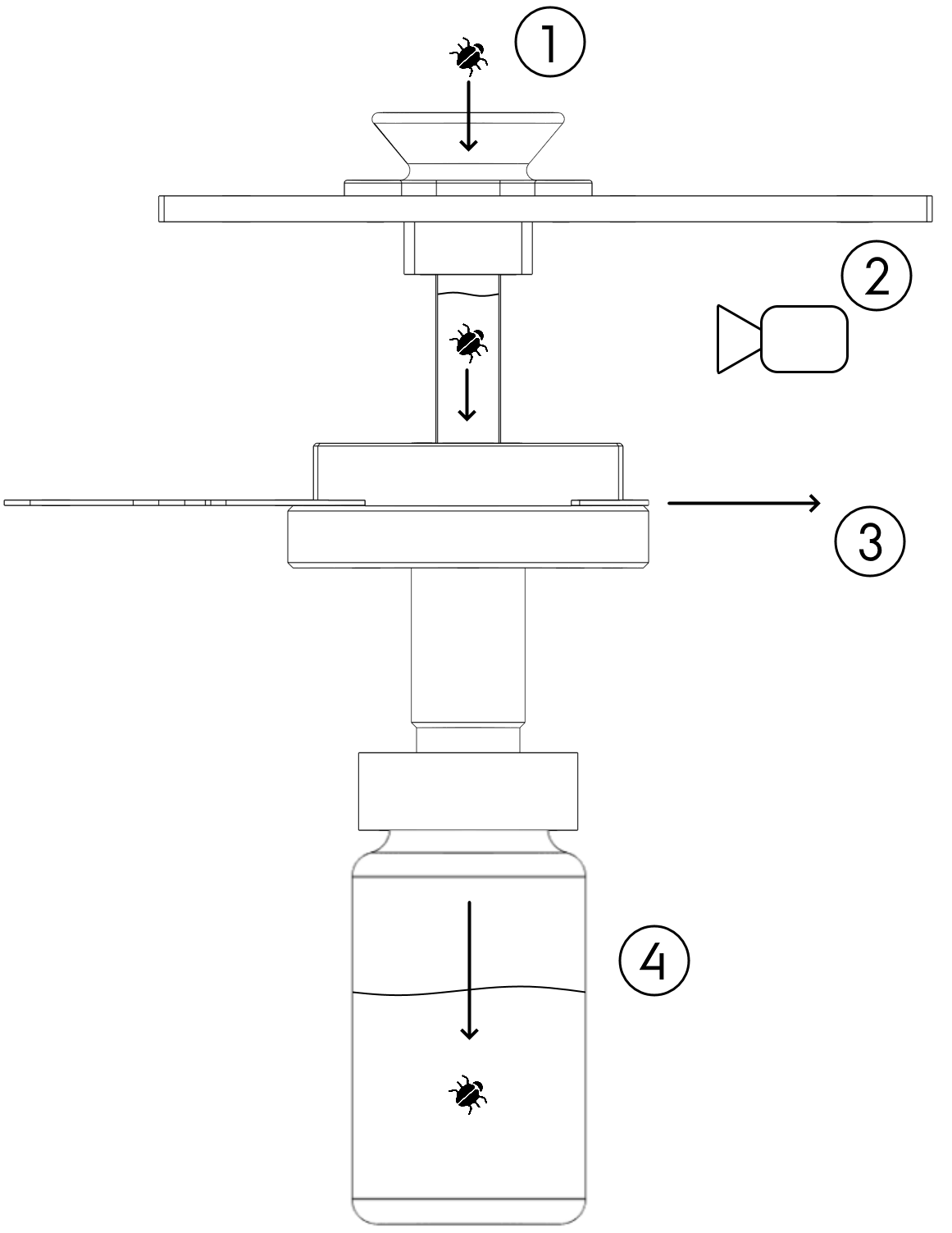}
	\caption{The flush through system of the BIODISCOVER machine. (1) A funnel helps filling the cuvette with ethanol without airbubbles, (2) as the specimen floats in ethanol, two cameras capture images of it from two angles, (3) a valve is opened to flush the specimen through to (4) a container for further archiving.}
	\label{system}
\end{figure}

The BIODISCOVER machine enables imaging multiple specimens before having to be emptied and refilling the cuvette. This is accommodated by a small area at the bottom of the cuvette, where the specimens are outside of the field of view of the cameras. Once a sample is imaged, the software triggers the opening of a sliding plate, which acts as a valve and flushes the specimens into a container below the imaging device case. Several containers placed in a rack can be controlled by the software based on input from the classification algorithm used to identify species. This enables a sorting of specimens into predefined classes based on size or taxonomy. In this way, the system can, for instance, separate large and small specimens for further molecular study, separate insect orders, or separate common and rare species. The system is described in Fig. \ref{system}. After the specimens have been flushed into the container for archiving, the pump in Fig. \ref{outside} is used to refill the cuvette with ethanol.

\begin{figure}[htbp]
	\centering
	\subfloat[]{
		\includegraphics[width=8cm]{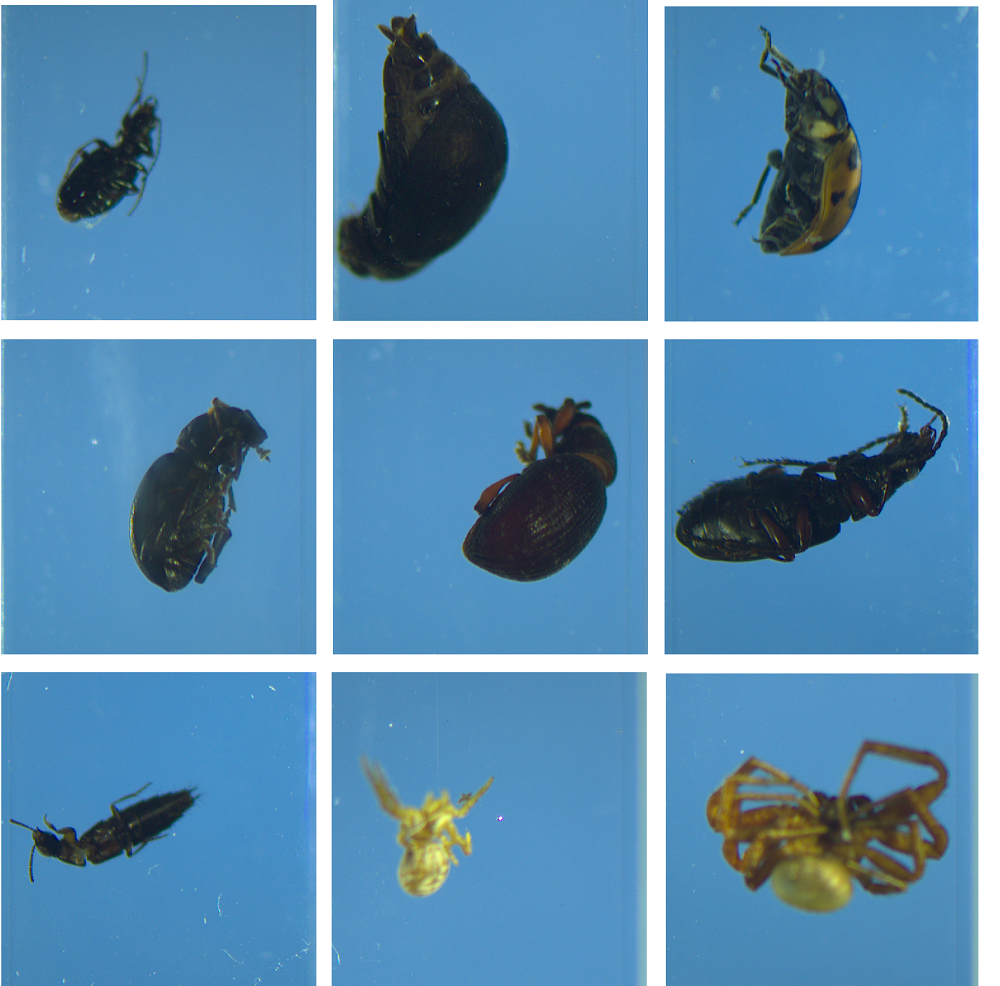}
		\label{all_taxa}
	}
	\subfloat[]{
		\includegraphics[width=8cm]{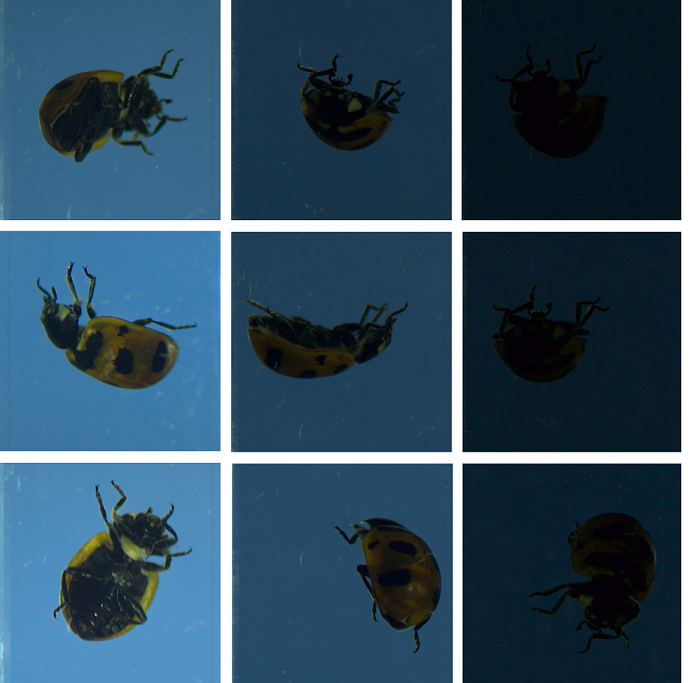}
		\label{coc_tra}
	}
	\caption{(a) The 9 species included in the image dataset. From top left: \textit{Bembidion grapii}, \textit{Byrrhus fasciatus}, \textit{Coccinella transversoguttata}, \textit{Otiorhynchus arcticus}, \textit{Otiorhynchus nodosus}, \textit{Patrobus septentrionus}, \textit{Quedius fellmanni}, \textit{Xysticus deichmanni} and \textit{Xysticus durus}. (b) Example images of a \textit{Coccinella transversoguttata} specimen with different camera settings using the BIODISCOVER machine. The exposure setting goes from top to bottom $\left[1000, 1500, 2000\right]$ and the aperture from left to right $\left[3.8, 8, 16\right]$.}
	\label{all_taxa_and_settings}
\end{figure}

Prior to large-scale imaging of reference collections of specimens of known identity, it is important to test the camera settings. As we plan to use the BIODISCOVER machine to create a large image database covering both terrestrial and aquatic invertebrates, it is important to optimize the different settings of the device to ensure the best possible image quality  of the database with regards to classification accuracy. For this purpose, we imaged a pilot dataset with nine different combinations of camera settings. To study the importance of lighting, we explored the effect of varying exposure values of $\left[1000 \mu\text{s}, 1500 \mu\text{s}, 2000 \mu\text{s}\right]$ and to study the effect of the focal length, we explored aperture values $\left[1:3.8,1:8,1:16\right]$. Using the nine different combinations of camera settings, we imaged a dataset of nine terrestrial arthropod species collected at Narsarsuaq, South Greenland and identified by morphology using \citet{bocher2015}: \textit{Bembidion grapii}, \textit{Byrrhus fasciatus}, \textit{Coccinella transversoguttata}, \textit{Otiorhynchus arcticus}, \textit{Otiorhynchus nodosus}, \textit{Patrobus septentrionus}, \textit{Quedius fellmanni}, \textit{Xysticus deichmanni} and \textit{Xysticus durus} (see Fig. \ref{all_taxa}). For the pilot data we wanted to include both species that have clear visual differences and should be easily indentifiable and species from the same genera that have similar morphological features and are more difficult to tell apart. 

The resulting nine datasets include the same specimens but the number of images varies depending on the camera settings since longer exposure time decreases the frame rate. Fig. \ref{coc_tra} shows example images of the same \textit{Coccinella transversoguttata} specimen from each of the nine camera setting combinations. Some of the specimens were damaged during the imaging. Therefore to have comparable results, we removed any specimens that were not present in all nine datasets. In addition, we performed a crude, initial check for outliers by calculating the mean of blue, green and red pixel values per species and making a list of all specimens that had mean pixel values further than three standard deviations from the species average. We then manually checked the images of those listed specimens and removed, e.g., images with only air bubbles or severed limbs. After this initial check, the number of images per specimen in the final data ranged from 1 to 376 (with 15 cases where a specimen had only 1 image). Table \ref{statistics} gives details on the final data.

\begin{table}[htbp]
	\caption{Image data details. This table gives the number of images in each dataset imaged with different camera setting (exposure $=\left[1000\mu\text{s},1500\mu\text{s},2000\mu\text{s}\right]$ and aperture $=\left[1:3.8,1:8,1:16\right]$). The number of specimens is the same for all datasets.}
	\begin{center}
		\begin{tabular}{|l|r|r|r|r|r|r|r|r|r|r|}
			\hline
			 & & \multicolumn{9}{c|}{\#Images} \\
			\hline
			 & & 1000 & 1000 & 1000 & 1500 & 1500 & 1500 & 2000 & 2000 & 2000 \\
			Species & \#Specimens & 1:3.8 & 1:8 & 1:16 & 1:3.8 & 1:8 & 1:16 & 1:3.8 & 1:8 & 1:16 \\
			\hline
			\textit{Bembidion grapii} & 17 & 2274 & 2554 & 2619 & 1677 & 1625 & 1741 & 1268 & 1270 & 1266 \\
			\hline
			\textit{Byrrhus fasciatus} & 52 & 4344 & 4778 & 5157 & 3222 & 3402 & 3054 & 2371 & 2262 & 2278 \\
			\hline
			\textit{Coccinella transversoguttata} & 57 & 5705 & 5607 & 5770 & 3958 & 3962 & 4183 & 2776 & 2770 & 2748 \\
			\hline
			\textit{Otiorhynchus arcticus} & 50 & 3197 & 3318 & 3488 & 2556 & 2220 & 2225 & 1898 & 1614 & 1700 \\
			\hline
			\textit{Otiorhynchus nodosus} & 139 & 9166 & 10010 & 9864 & 6818 & 6524 & 6571 & 4796 & 4563 & 4690 \\
			\hline
			\textit{Patrobus septentrionus} & 108 & 11056 & 11583 & 11738 & 8383 & 8311 & 8028 & 6004 & 5808 & 6148 \\
			\hline
			\textit{Quedius fellmanni} & 42 & 5749 & 6438 & 6363 & 4577 & 4461 & 4393 & 3708 & 3270 & 3318 \\
			\hline
			\textit{Xysticus deichmanni} & 25 & 2434 & 2709 & 2611 & 1800 & 1890 & 1802 & 1680 & 1363 & 1364 \\
			\hline
			\textit{Xysticus durus} & 43 & 3997 & 4113 & 4043 & 3036 & 2922 & 2841 & 2212 & 2119 & 2110 \\
			\hline
		\end{tabular}
	\end{center}
	\label{statistics}
\end{table}

We split the data into training ($70\%$), validation ($10\%$) and test ($20\%$) observations. As difficult specimens can introduce variation to the results, we performed the tests on 10 different random data divisions. If a specimen was selected for training, all the images of that specimen were used for training. To keep the results comparable between the different camera settings, we used the exact same training specimens for all camera settings. Respectively, the exact same validation and testing specimens were used for each camera setting combination. The number of images, exposure and aperture differed for the camera setting combinations but the specimens remained the same, i.e. if a difficult, atypical specimen of a certain species was selected for testing that same specimen was used for testing all the camera setting combinations, making the identification task equally difficult for all the settings.

To examine whether the BIODISCOVER machine benefits from having two cameras shooting from different angles, we performed a test where, for each specimen, we counted the number of images captured by each of the cameras. To compare the two camera angles, we require an equal amount of images from both cameras. For each specimen, we checked which camera had captured less images and randomly sampled the same amount of images from the other camera as well. Finally, we randomly sampled that same amount of images for each specimen, this time including images from both cameras. Thus, we obtained three datasets, each with the same total amount of images. To account for variation in a single data split, we ran the test again on 10 data divisions into training, validation and test observations.

For the classification task, we tested two widely used deep \acp{CNN}, namingly Resnet-50 \citep{he2016} and InceptionV3 \citep{szegedy2016}, both pretrained with the Imagenet database \citep{deng2009}. For each data division, we used the training observations to fine-tune the weights of the pre-trained \acp{CNN}. In order to feed the images to the network, we scaled them all to 128$\times$128 pixels. This caused slight distortion to specimens taller than 496 px but the majority of the images ($86\%$) are square-shaped and thus remained undistorted. We used batch normalization, a batch size of 128, and a decaying learning rate $\left[0.001, 0.0001,0.00001, 0.000001\right]$, training the network for 50 epochs with each learning rate. The validation images were used to select optimal weights for the network. Finally, the test observations were used to test the final classification accuracy.

As we used multiple images per observation, we needed to define a decision rule to determine the final species of the observation based on the predictions for all the images. The simplest option was to use majority vote, i.e. the species that was predicted most often among the images of the specimen was chosen as the final prediction. 

The BIODISCOVER machine derives geometric features from each image taken of each individual. These features include the area of the specimen in the image, which can be used for biomass prediction. For this purpose, we imaged three species of \textit{Diptera} with the optimal camera settings and measured dry weight for a subset of this data ($n=65$). The species included in this data set were \textit{Dolichopus groenlandicus}, \textit{Dolichopus plumipes} and \textit{Tachina ampliforceps}. The area was calculated from images as average per specimen. After imaging, each specimen was dried at 70°C for 48 hours and weighed on a scale to the nearest 0.0001g to quantify dry weight. For biomass prediction, we performed a logarithm transformation on the data and fitted a linear mixed model  to examine the relationship between the average area and dry weight, using the species as a random factor. However, the model assumptions could not be met with the data, hence, we fitted separate generalized linear models for each species. 

\section{Results}

\begin{figure}[htbp]
	\centering
	\subfloat[]{
		\includegraphics[width=9cm]{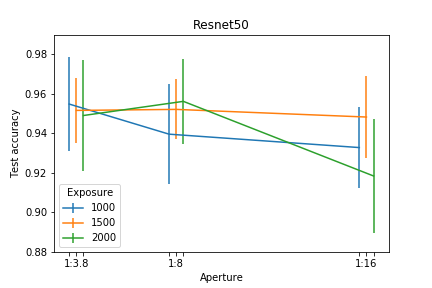}
		\label{acc_resnet}
	}
	\subfloat[]{
		\includegraphics[width=9cm]{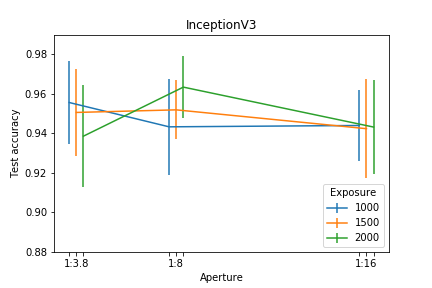}
		\label{acc_iv3}
	}\\
	\caption{Average test classification accuracies $\pm$ standard deviation for different camera setting combinations for (a) Resnet-50 network, (b) InceptionV3 network.}
	\label{camera_results_fig}
\end{figure}

Our first objective was to find optimal camera settings for the imaging device for species identification. The average classification accuracy across 10 test sets is presented in Table \ref{camera_results} and Fig. \ref{camera_results_fig}. Based on the results from our pilot data, the optimal camera settings for both \acp{CNN} were exposure $=2000\mu\text{s}$ and aperture $=1:8$. The InceptionV3 network produced the highest classification accuracy with these camera settings. For InceptionV3, the best camera settings also yielded the second lowest standard deviation. The differences between the settings were small but we observed that decreasing aperture to 1:16 decreased the classification accuracy. For higher exposure, an initial decrease in aperture enhanced the results while decreasing aperture to 1:16 decreased classification accuracy. For exposure $=1000\mu\text{s}$, even increasing aperture to 8 decreased classification accuracy. The optimal camera settings are intuitive as they provide sharp images while having as much light as possible. 

\begin{table}[htbp]
	\caption{Average classification accuracy (and standard deviation) of test specimens for nine camera setting combinations (exposure $=\left[1000\mu\text{s},1500\mu\text{s},2000\mu\text{s}\right]$ and aperture $=\left[1:3.8,1:8,1:16\right]$). The highest classification accuracy is marked in bold. The final class is decided by majority rule.}
	\begin{center}
		\begin{tabular}{|c|rr|rr|rr|rr|rr|rr|}
			\hline
			& \multicolumn{6}{c|}{Resnet-50} & \multicolumn{6}{c|}{InceptionV3}\\
			& \multicolumn{2}{c}{1000} & \multicolumn{2}{c}{1500} & \multicolumn{2}{c|}{2000} 	& \multicolumn{2}{c}{1000} & \multicolumn{2}{c}{1500} & \multicolumn{2}{c|}{2000}\\
			\hline
			1:3.8 & 0.955 & (0.024) & 0.952 & (0.016) & 0.949 & (0.028) & 0.956 & (0.021) & 0.951 & (0.022) & 0.939 & (0.026)\\
			1:8 & 0.940 & (0.025) & 0.952 & (0.015) & 0.956 & (0.022) & 0.943 & (0.024) & 0.952 & (0.015) & \textbf{0.963} & (0.016)\\
			1:16 & 0.933 & (0.020) & 0.948 & (0.021) & 0.918 & (0.029) & 0.944 & (0.018) & 0.942 & (0.025) & 0.943 & (0.024)\\
			\hline
		\end{tabular}
	\end{center}
	\label{camera_results}
\end{table}

In addition to majority vote, we used the weighted sum rule for \ac{CNN} confidence values presented in \citet{raitoharju2019b}. Using the weighted sum rule for confidence values for our image data gave varying results. For Resnet-50, the weighted sum rule produced slightly higher accuracies across all camera settings but for InceptionV3, it produced sometimes lower and sometimes higher accuracies with the highest value for exposure $=2000\mu\text{s}$, aperture $=1:8$ being almost identical to that of the majority vote rule. Hence, we used the majority vote decision rule in our classification results. The results for the weighted sum rule are shown in detail in Table \ref{weighted_sum} in the Appendix.

Table \ref{camera_time} gives the average training times (on K80 GPU) for the different camera settings. The differences in training time were mainly due to the number of images (see Table \ref{statistics}), but confirmed that the optimal camera settings were optimal with regards to the training time as well. In addition to producing higher classification accuracy, InceptionV3 network was also faster to train.

\begin{table}[htbp]
	\caption{Average training time (hh:mm:ss) of \acp{CNN} for nine camera setting combinations (exposure $=\left[1000\mu\text{s},1500\mu\text{s},2000\mu\text{s}\right]$ and aperture $=\left[1:3.8,1:8,1:16\right]$).}
	\begin{center}
		\begin{tabular}{|c|r|r|r|r|r|r|}
			\hline
			& \multicolumn{3}{c|}{Resnet-50} & \multicolumn{3}{c|}{InceptionV3}\\
			\cline{2-7}
			& 1000 & 1500 & 2000 & 1000 & 1500 & 2000\\
			\hline
			1:3.8 & 28:37:52 & 21:52:37  & 15:58:08  & 26:31:28 & 19:44:05 & 14:39:07\\
			1:8 & 31:03:12  & 21:11:10  & 15:08:08  & 28:41:15 & 18:50:00 & 13:55:02\\
			1:16 & 31:19:24  & 20:56:28  & 15:13:01  & 28:57:44 & 18:37:59 & 14:08:15\\
			\hline
		\end{tabular}
	\end{center}
	\label{camera_time}
\end{table}

To test whether the BIODISCOVER machine benefits from having two cameras shooting from different angles, we performed a test on the data imaged with the optimal camera settings. The results are shown in Table \ref{camera_angle}. The classification accuracy was higher when using images from both cameras. In addition, for Resnet-50, the standard deviation was lower meaning there is less variation in the classification accuracy due to choice of test specimens. The classification accuracies in this test are slightly lower than in Table \ref{camera_results} as for this particular test we are using less images per specimen (approximately 50\%).

\begin{table}[htbp]
	\caption{Mean and standard deviation of classification accuracy of test specimens over 10 data splits using images from single cameras vs. both cameras. The camera settings for the data are exposure $=2000\mu\text{s}$ and aperture $=1:8$. The number of images per set is fixed.}
	\begin{center}
		\begin{tabular}{|l|r|r|r|r|r|r|}
			\hline
			& \multicolumn{3}{c|}{Resnet-50} & \multicolumn{3}{c|}{InceptionV3}\\
			\cline{2-7}
			& Camera 1 & Camera 2 & Both cameras & Camera 1 & Camera 2 & Both cameras\\
			\cline{2-7}
			$\overline{Acc}_{test}$ & 0.946 & 0.943 & 0.955  & 0.950 & 0.944 & 0.952\\
			$std(Acc_{test})$ & 0.023 & 0.024 & 0.018 & 0.019 & 0.029 & 0.026\\
			\hline
		\end{tabular}
	\end{center}
	\label{camera_angle}
\end{table}

Once we had optimized the camera settings, we re-ran the InceptionV3 network with the data including also the three \textit{Diptera} species. The average classification accuracy over 10 test sets was 0.980. The information of individual classification decisions is shown in a confusion matrix with the true species on the rows and the predicted species on the columns. Table \ref{conf} shows the normalized average confusion matrix over the 10 random data splits for InceptionV3 \ac{CNN} with the optimal camera settings. As for individual species, \textit{Bembidion grapii} was the hardest to identify. Some of the specimens were misclassified as \textit{Patrobus septentrionus} and \textit{Quedius fellmanni}. In addition, \textit{Otiorhyncus arcticus} and \textit{Otiorhynchus nodosus} were often confused, as well as \textit{Xysticus deichmanni} and \textit{Xysticus durus}. Other common classification errors were  misclassifying \textit{Byrrhus fasciatus} as \textit{Otiorhynchus nodosus} and misclassifying \textit{Xysticus durus} as \textit{Bembidion grapii}. The species that performed poorly compared to the others are species with the lowest number of images in the data. The accuracy could be improved with collecting more data on these species or by using data augmentation techniques.

\begin{table}[htbp]
	\caption{Normalized average confusion matrix over 10 random data split for data imaged with exposure $=2000\mu\text{s}$ and aperture $=1:8$, classified with InceptionV3 network. The rows of the table represent the true species while the columns represent the predicted species and the cells give the average percentage over the 10 test data.}
	\begin{center}
		\begin{tabular}{|l|r|r|r|r|r|r|r|r|r|r|r|r|}
			\hline
			& \textit{Be\_gr} & \textit{By\_fa} & \textit{Co\_tr} & \textit{Do\_gr} & \textit{Do\_pl} & \textit{Ot\_ar} & \textit{Ot\_no} & \textit{Pa\_se} & \textit{Qu\_fel} & \textit{Ta\_am} &\textit{Xy\_de} & \textit{Xy\_du}\\
			\hline
			\textit{Be\_gr} & \textbf{0.756} & 0 & 0 & 0 & 0 & 0 & 0 & 0.122 & 0.122 & 0 & 0 & 0\\
			\textit{By\_fa} & 0 & \textbf{0.992} & 0 & 0 & 0 & 0 & 0.008 & 0 & 0 & 0 & 0 & 0\\
			\textit{Co\_tr} & 0 & 0 & \textbf{1.0} & 0 & 0 & 0 & 0 & 0 & 0 & 0 & 0 & 0\\
			\textit{Do\_gr} & 0 & 0 & 0 & \textbf{1.0} & 0 & 0 & 0 & 0 & 0 & 0 & 0 & 0\\ 
			\textit{Do\_pl} & 0 & 0 & 0 & 0.015 & \textbf{0.985} & 0 & 0 & 0 & 0 & 0 & 0 & 0\\
			\textit{Ot\_ar} & 0 & 0 & 0 & 0 & 0 & \textbf{0.910} & 0.090 & 0 & 0 & 0 & 0 & 0\\
			\textit{Ot\_no} & 0 & 0.004 & 0 & 0 & 0 & 0.019 & \textbf{0.977} & 0 & 0 & 0 & 0 & 0\\
			\textit{Pa\_se} & 0 & 0 & 0 & 0 & 0 & 0.004 & 0.004 & \textbf{0.991} & 0 & 0 & 0 & 0\\
			\textit{Qu\_fel} & 0 & 0 & 0.010 & 0 & 0.010 & 0 & 0 & 0 & \textbf{0.980} & 0 & 0 & 0\\
			\textit{Ta\_am} & 0 & 0 & 0 & 0.005 & 0 & 0 & 0 & 0.010 & 0 & \textbf{0.985} & 0 & 0\\
			\textit{Xy\_de} & 0 & 0 & 0 & 0 & 0 & 0 & 0 & 0 & 0 & 0 & \textbf{0.941} & 0.059\\
			\textit{Xy\_du} & 0.029 & 0 & 0 & 0 & 0 & 0 & 0 & 0 & 0.010 & 0 & 0.029 & \textbf{0.933}\\
			\hline
		\end{tabular}
	\end{center}
	\label{conf}
\end{table}



\begin{figure}[htbp]
	\centering
	\includegraphics[width=9cm]{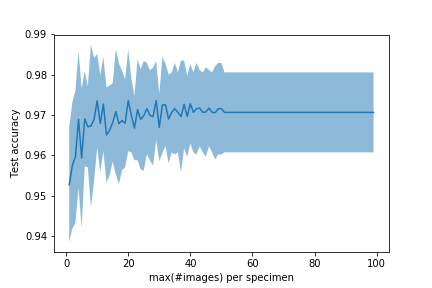}
	\caption{Classification accuracy of the test specimens plotted against the maximum number of images per specimen. The solid line shows the average over 10 data divisions and the light blue area represents the average $\pm$ standard deviation.}
	\label{num_images}
\end{figure}

When considering automated biomonitoring, one key factor is the time it takes to automatically identify the taxonomic identity of a specimen. Training of the network can take a long time but it needs to be done only once so we recommend to use as much data as possible for the training. In taxa identification scenarios, optimising the time used for testing is more interesting. The number of images per specimen affects the total time of identification as each image needs a prediction. To optimize the number of images per specimen, we tested how this affects the classification accuracy. As the specimens had varying number of images, we tested with the maximum number of images per specimen, $N_{max}$. If a specimen had less images, we used all of them. If a specimen had more images, we randomly sampled $N_{max}$ of them. Again, we ran this test on the 10 data splits imaged with the optimal camera settings. The results are shown in Fig. \ref{num_images}, where the dark blue line represents the average classification accuracy over 10 data splits and the lighter blue area is $\pm$ standard deviation. The average number of images per specimen is 47 so while some specimens had over 100 images, the test accuracy stabilized at approximately 50 images. The same accuracy of approximately 96\% could already be achieved with 20 images per specimen but lower numbers of images increased the variation in the classification accuracy. While increasing the maximum number of images per specimen does increase the time for taxa predictions, testing time is not an issue. Even with a maximum 100 images per specimen the time taken to predict taxa for the entire test data was on average 40 seconds. However, fixing the maximum number of images per specimen would mean less images for the BIODISCOVER device to store onto the computer, enabling faster imaging process and saving computational resources.

\begin{figure}[htbp]
	\centering
	\includegraphics[width=17cm]{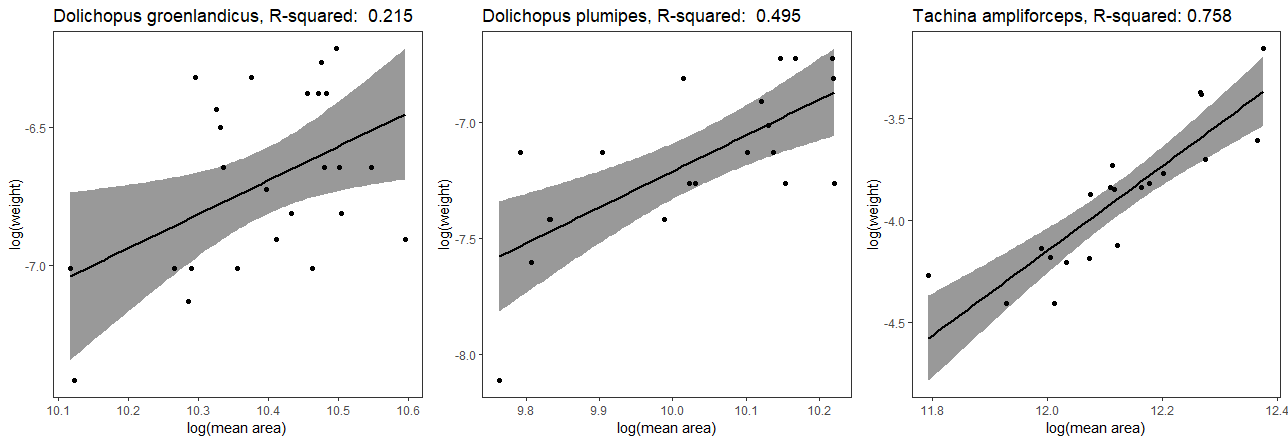}
	\caption{Generalized linear models for biomass of \textit{Dolichopus groenlandicus}, \textit{Dolichopus plumipes} and \textit{Tachina ampliforceps} with average area predicting dry weight.}
	\label{biomass}
\end{figure}

Fig. \ref{biomass} shows the results of the biomass prediction. The logarithm transformed average area was found to be statistically significant predictor of dry weight for all three \textit{Diptera} species. However, considering the R-squares of the different models, the average area is a good predictor only for the largest species, \textit{Tachina ampliforceps} (r-squared = 0.758). For the two small Dolichopus species, relationships were weaker.

\section{Discussion}

We have presented an image-based identification system (i.e. the BIODISCOVER machine) for insects and other invertebrates as an alternative to manual identification. We demonstrated a very high classification accuracy on a test set of images of 249 specimens of known identity belonging to one of 12 insect and spider species. We were also able to show that biomass of individual specimens could be predicted straight from information in the images. Together, our results pave the way for future non-destructive, automatic, image-based identification and biomass estimation of bulk invertebrate samples.

We imaged specimens of seven beetle, two spider, and three fly species with the BIODISCOVER machine with different values for exposure time and aperture settings and found that the best camera settings were obtained with an exposure time $2000\mu s$ and an aperture $1:8$. With these settings, we obtained a high test classification accuracy of 98.0\%, demonstrating the great potential of the BIODISCOVER machine for the use in species identification. In \citet{arje2020}, e.g., taxonomic experts achieved an accuracy of 93.9\% with a dataset of 39 taxonomic groups. While adding more species to the data will increase the difficulty of the classification task \citep{arje2020}, data augmentation can be used to improve the results for rare species \citep{raitoharju2016}.

We tested predicting biomass from images on a subset of three fly species. We explored a joined mixed model for all species but the small data restricted our final analysis to three species-wise generalized linear models. The average area of the specimen was a good predictor for dry weight for the largest species, \textit{Tachina ampliforceps}, but the two smaller species would require more data for better results. For instance, by weighing more species of different sizes, it would be possible to quantify the uncertainty associated with using general relationships between area and dry weight constructed from multiple, related species (e.g. species belonging to the same family). The BIODISCOVER machine can easily be used with any animal small enough to fit into the cuvette.  Since the imaging device comprises standard industry components, ensuring the possibility to build more copies of the BIODISCOVER machine. The flow-through and refill systems facilitate easy archiving of samples. Furthermore, the BIODISCOVER machine also saves metadata from the images, e.g. geometric features that can be used in automatic biomass predictions. 

The imaging device is one of three components for automatic image-based species identification. We are currently working on implementing a) a computer-vision enabled robotic arm to automatically detect insects from a bulk sample in a tray, choose among different tools to move individual specimens to the imaging device and b) a sorting rack to place specimens in the preferred container after imaging based on e.g. taxonomic identity, size or rarity. With these additions, the BIODISCOVER machine offers high throughput, non-destructive taxonomic identification, size/biomass estimation, counting and further morphological data, while keeping the specimens intact. Given that the robotic arm is standard industry equipment, we are on the verge of producing a truly automated species identification system for invertebrates, both aquatic and terrestrial.

\section{Acknowledgements}
We would like to thank CSC for computational resources. TTH acknowledges funding from his VILLUM Experiment project “Automatic Insect Detection” (grant 17523) and an Aarhus University synergy grant. 

\section{Authors' contributions}

JÄ drafted the paper with contributions from TTH with the other authors providing feedback and approving the final manuscript. CM, TTH, MRJ and SAM designed and built the BIODISCOVER machine with inputs from KM and VT who had designed the prototype described in \citet{raitoharju2018}. MSR imaged the arthropod specimens and JÄ, TTH, AI, MG and JR designed the classification experiments for the data.

\section*{Data availability}
The image data will be made publicly available on publication of the manuscript.

\bibliography{references}

\begin{thebibliography}{}

\bibitem[\protect\astroncite{\"Arje et~al.}{subm}]{arje2020}
\"Arje, J., J.~Raitoharju, A.~Iosifidis, V.~Tirronen, K.~Meissner, M.~Gabbouj,
  S.~Kiranyaz, and S.~K\"arkk\"ainen\leavevmode\nopagebreak\newline subm.
\newblock Human experts vs. machines in taxa recognition.
\newblock {\em (submitted)}.

\bibitem[\protect\astroncite{Aylagas et~al.}{2016}]{aylagas2016}
Aylagas, E., A.~Borja, X.~Irigoien, and
  N.~Rodríguez-Ezpeleta\leavevmode\nopagebreak\newline 2016.
\newblock Benchmarking {DNA} metabarcoding for biodiversity-based monitoring
  and assessment.
\newblock {\em Frontiers in Marine Science}, 3:1809--12.

\bibitem[\protect\astroncite{B\"ocher et~al.}{2015}]{bocher2015}
B\"ocher, J., N.~P. Kristensen, T.~Pape, and L.~V.
  (Eds.)\leavevmode\nopagebreak\newline 2015.
\newblock {\em The {Greenland} entomofauna: an identification manual of
  insects, spiders and their allies}.
\newblock Brill.

\bibitem[\protect\astroncite{Bochinski et~al.}{2018}]{bochinski2018}
Bochinski, E., G.~Bacha, V.~Eiselein, T.~J.~W. Walles, J.~C. Nejstgaard, and
  T.~Sikora\leavevmode\nopagebreak\newline 2018.
\newblock Deep active learning for in situ plankton classification.
\newblock {\em International Conference on Pattern Recognition, Springer}, Pp.~
  5--15.

\bibitem[\protect\astroncite{Borja and Elliott}{2013}]{borja2013}
Borja, A. and M.~Elliott\leavevmode\nopagebreak\newline 2013.
\newblock Marine monitoring during an economic crisis: the cure is worse than
  the disease.
\newblock {\em Marine Pollution Bulletin}, 68:1--3.

\bibitem[\protect\astroncite{Dai et~al.}{2016}]{dai2016}
Dai, J., R.~Wang, H.~Zheng, G.~Ju, and X.~Qiao\leavevmode\nopagebreak\newline
  2016.
\newblock {ZooplankoNet}: deep convolutional network for zooplankton
  classification.
\newblock {\em OCEANS 2016, IEEE}, Pp.~ 1--6.

\bibitem[\protect\astroncite{Deng et~al.}{2009}]{deng2009}
Deng, J., W.~Dong, R.~Socher, L.~Li, K.~Li, and
  L.~{Fei-Fei}\leavevmode\nopagebreak\newline 2009.
\newblock {ImageNet:} a large-sacel hierarchical image database.
\newblock {\em IEEE Computer Vision and Pattern Recognition (CVPR)}.

\bibitem[\protect\astroncite{Dunker et~al.}{2016}]{dunker2016}
Dunker, K.~J., A.~J. Spulveda, R.~L. Massengill, J.~B. Olsen, O.~L. Russ, J.~K.
  Wenburg, and A.~Antonovich\leavevmode\nopagebreak\newline 2016.
\newblock Potential of environmental {DNA} to evaluate northern pike ({E}sox
  lucius) eradication efforts: an experimental test and case study.
\newblock {\em PLos ONE}, 11(9):e0162277.

\bibitem[\protect\astroncite{Elbrecht et~al.}{2017}]{elbrecht2017}
Elbrecht, V., E.~E. Vamos, K.~Meissner, J.~Aroviita, and
  F.~Leese\leavevmode\nopagebreak\newline 2017.
\newblock Assessing strengths and weaknesses of dna metabarcoding-based
  macroinvertebrate identification for routine stream monitoring.
\newblock {\em Methdos in Ecology and Evolution}, 8(10):1265--1275.

\bibitem[\protect\astroncite{Feng et~al.}{2016}]{feng2016}
Feng, L., B.~Bhanu, and J.~Heraty\leavevmode\nopagebreak\newline 2016.
\newblock A software system for automated identification and retrieval of moth
  images based on wing attributes.
\newblock {\em Pattern Recognition}, 51:225--241.

\bibitem[\protect\astroncite{Gaston and O'Neill}{2004}]{gaston2004}
Gaston, K.~J. and M.~A. O'Neill\leavevmode\nopagebreak\newline 2004.
\newblock Automates species identification: why not?
\newblock {\em Philosophical Transactions of the Royal Society of London.
  Series B: Biological Scienses}, 359(1444):655--667.

\bibitem[\protect\astroncite{Hallmann et~al.}{2017}]{hallmann2017}
Hallmann, C.~A., M.~Sorg, E.~Jongejans, H.~Siepel, N.~Hofland, H.~Schwan,
  W.~Stenmans, A.~M\"uller, H.~Sumser, T.~H\"orren, D.~Goulson, and H.~{de
  Kroon}\leavevmode\nopagebreak\newline 2017.
\newblock More than 75 percent decline over 27 years in total flying insect
  biomass in protected areas.
\newblock {\em PLoS ONE}, 12(10).

\bibitem[\protect\astroncite{Hansen et~al.}{2016}]{hansen2016}
Hansen, R.~R., O.~L.~P. Hansen, J.~J. Bowden, U.~A. Treier, S.~Normand, and
  T.~T. H{\o}ye\leavevmode\nopagebreak\newline 2016.
\newblock Meter scale variation in shrub dominance and soil moisture structure
  {Arctic} arthropod communities.
\newblock {\em PeerJ}, 4:e2224.

\bibitem[\protect\astroncite{He et~al.}{2016}]{he2016}
He, K., X.~Zhang, S.~Ren, and J.~Sun\leavevmode\nopagebreak\newline 2016.
\newblock Deep residual learning for image recognition.
\newblock {\em The IEEE Conference on Computer Vision and Pattern Recognition
  (CVPR)}, Pp.~ 770--778.

\bibitem[\protect\astroncite{Hortal et~al.}{2015}]{hortal2015}
Hortal, J., F.~{de Bello}, J.~A.~F. {Diniz-Filho}, T.~M. Lewinsohn, J.~M. Logo,
  and R.~J. Ladle\leavevmode\nopagebreak\newline 2015.
\newblock Seven shortfalls that beset large-scale knowledge of biodiversity.
\newblock {\em Annual Review of Ecology, Evolution, and Systematics},
  46(1):523--549.

\bibitem[\protect\astroncite{H{\o}ye and Forchhammer}{2008}]{hoye2008}
H{\o}ye, T.~T. and M.~C. Forchhammer\leavevmode\nopagebreak\newline 2008.
\newblock Phenology of high-arctic arthropods: effects of climate on spatial,
  seasonal and inter-annual variation.
\newblock {\em Insect Conservation and Diversity}, 40:299--324.

\bibitem[\protect\astroncite{Kermarrec et~al.}{2014}]{kermarrec2014}
Kermarrec, L., A.~Franc, F.~Rimet, P.~Chaumeil, J.-M. Frigerio, J.-F. Humbert,
  and A.~Bouchez\leavevmode\nopagebreak\newline 2014.
\newblock A next-generation sequencing approach to river biomonitoring using
  benthic diatoms.
\newblock {\em Freshwater Science}, 33:349--363.

\bibitem[\protect\astroncite{Keskin}{2014}]{keskin2014}
Keskin, E.\leavevmode\nopagebreak\newline 2014.
\newblock Detection of invasive freswater fish species using evironmental dna
  survey.
\newblock {\em Biochemical Systematics and Ecology}, 56:68--74.

\bibitem[\protect\astroncite{LeQuing and Zhen}{2012}]{lequing2012}
LeQuing, Z. and Z.~Zhen\leavevmode\nopagebreak\newline 2012.
\newblock Automatic insect classification based on local mean colour feature
  and supported vector machines.
\newblock {\em Oriental Insects}, 46(3/4):260--269.

\bibitem[\protect\astroncite{Liu et~al.}{2008}]{liu2008}
Liu, F., S.~Z.-R, J.-W. Zhang, and H.-Z. Yang\leavevmode\nopagebreak\newline
  2008.
\newblock Automatic insect identification based on color characters.
\newblock {\em Chinese Bulletin of Entomology}, 45:150--153.

\bibitem[\protect\astroncite{Loboda et~al.}{2017}]{loboda2017}
Loboda, S., J.~Savage, C.~M. Buddle, N.~M. Schmidt, and T.~T.
  H{\o}ye\leavevmode\nopagebreak\newline 2017.
\newblock Declining diversity and abundance of {High Arctic} fly assemblages
  over two decades of rapid climate warming.
\newblock {\em Ecography}, 41:265--277.

\bibitem[\protect\astroncite{Nyg{\aa}rd et~al.}{2016}]{nygard2016}
Nyg{\aa}rd, H., S.~Oinonen, M.~Lehtiniemi, H.~H\"allfors, E.~Rantaj\"arvi, and
  L.~Uusitalo\leavevmode\nopagebreak\newline 2016.
\newblock Price versus value of marine monitoring.
\newblock {\em Frontiers in Marine Science}, 3:205.

\bibitem[\protect\astroncite{Perre et~al.}{2016}]{perre2016}
Perre, P., F.~A. Faria, L.~R. Jorge, A.~Rocha, R.~S. Torres, M.~F. Souza-Filho,
  T.~M. Lewinson, and R.~A. Zucchi\leavevmode\nopagebreak\newline 2016.
\newblock Toward and automated identification of \textit{Anastrepha} fruit
  flies in the \textit{fraterculus} group ({Diptera, Tephritidae}).
\newblock {\em Neotropical Entomology}, 45(5):554--558.

\bibitem[\protect\astroncite{Potamis}{2014}]{potamis2014}
Potamis, I.\leavevmode\nopagebreak\newline 2014.
\newblock Automatic classification of a taxon-rich communit recorded in the
  wild.
\newblock {\em PLOS ONE}, 9(5):e96936.

\bibitem[\protect\astroncite{Qian et~al.}{2011}]{qian2011}
Qian, L., W.~{HongBin}, Z.~Zhen, and
  K.~{XiangBo}\leavevmode\nopagebreak\newline 2011.
\newblock Automatic stridulation identification of bark beetles based on {MFCC}
  and {BP} network.
\newblock {\em Jourlan of Beijing Forestry University}, 33(5):81--85.

\bibitem[\protect\astroncite{Raitoharju and Meissner}{2019}]{raitoharju2019b}
Raitoharju, J. and K.~Meissner\leavevmode\nopagebreak\newline 2019.
\newblock On confidences and their use in (semi-)automatic multi-image taxa
  identification.
\newblock {\em Proceeding of IEEE Symposium Series on Computational
  Intelligence}.

\bibitem[\protect\astroncite{Raitoharju et~al.}{2018}]{raitoharju2018}
Raitoharju, J., E.~Riabchenko, I.~Ahmad, A.~Iosifidis, M.~Gabbouj, S.~Kiranyaz,
  V.~Tirronen, J.~\"Arje, S.~K\"arkk\"ainen, and
  K.~Meissner\leavevmode\nopagebreak\newline 2018.
\newblock Benchmark database for fine-grained image classification of benthic
  macroinverebrates.
\newblock {\em Image and Vision Computing}, 78:73--83.

\bibitem[\protect\astroncite{Raitoharju et~al.}{2016}]{raitoharju2016}
Raitoharju, J., E.~Riabchenko, K.~Meissner, I.~Ahmad, A.~Iosifidis, M.~Gabbouj,
  and S.~Kiranyaz\leavevmode\nopagebreak\newline 2016.
\newblock Data enrichment in fine-grained classification of aquatic
  macroinvertebrates.
\newblock {\em ICPR 2nd Workshop on Computer Vision for Analysis of Underwater
  Imagery (CVAUI)}.

\bibitem[\protect\astroncite{Raupach et~al.}{2010}]{raupach2010}
Raupach, M.~J., J.~J. Astrin, K.~Hannig, M.~K. Peters, M.~Y. Stoeckle, and
  J.-W. W\"agele\leavevmode\nopagebreak\newline 2010.
\newblock Molecular species identification of {Central European} ground beetles
  ({Coleoptera: Carabidae}) using nuclear {rDNA} expansion segments and {DNA}
  barcodes.
\newblock {\em Fronties in Zoology}, 7:26.

\bibitem[\protect\astroncite{Rich et~al.}{2013}]{rich2013}
Rich, M.~E., L.~Gough, and N.~T. Boelman\leavevmode\nopagebreak\newline 2013.
\newblock Arctic arthropod assemblages in habitats of differing shrub
  dominance.
\newblock {\em Ecography}, 36:994--1003.

\bibitem[\protect\astroncite{Santhi et~al.}{2013}]{santhi2013}
Santhi, N., C.~Pradeepa, P.~Subashini, and
  S.~Kalaiselvi\leavevmode\nopagebreak\newline 2013.
\newblock Automatic identification of algal community from microscopic images.
\newblock {\em Bioinformatics and Biology Insights}, 7:3.

\bibitem[\protect\astroncite{Schröder et~al.}{1995}]{schroder1995}
Schröder, S., W.~Drescher, V.~Steinhage, and
  B.~Kastenholz\leavevmode\nopagebreak\newline 1995.
\newblock An automated method for the identification of bee species
  ({Hymenoptera: Apoidea}).
\newblock {\em Proceedings of the International Symposium on Conserving
  Europe's Bees, International Bee Research Association \& Linnean Society}.

\bibitem[\protect\astroncite{Seibold et~al.}{2019}]{seibold2019}
Seibold, S., M.~M. Gossner, N.~K. Simons, N.~Bl\"utgen, J.~M\"uller,
  D.~Ambarli, C.~Ammer, J.~Bauhus, M.~Fischer, J.~C. Habel, K.~E. Linsenmair,
  T.~Nauss, C.~Peonen, D.~Prati, P.~Schall, E.-D. Schulze, J.~Vogt,
  S.~W\"ollauer, and W.~W. Weisser\leavevmode\nopagebreak\newline 2019.
\newblock Arthropod decline in grasslands and forests is associated with
  landscape-level drivers.
\newblock {\em Nature}, 574:671--674.

\bibitem[\protect\astroncite{Szegedy et~al.}{2016}]{szegedy2016}
Szegedy, C., V.~Canhoucke, S.~Ioffe, J.~Shlens, and
  Z.~Wojna\leavevmode\nopagebreak\newline 2016.
\newblock Rethink the inception architecture for computer vision.
\newblock {\em Proceedings of the IEEE conference on computer vision and
  pattern recognition}, Pp.~ 2818--2826.

\bibitem[\protect\astroncite{Timms et~al.}{2012}]{timms2012}
Timms, L.~L., J.~J. Bowden, K.~S. Summerville, and C.~M.
  Buddle\leavevmode\nopagebreak\newline 2012.
\newblock Does species-level resolution matter? {Taxonomic} sufficiency in
  terrestrial arthropod biodiversity studies.
\newblock {\em Insect Conservation and Diversity}, 6:453--462.

\bibitem[\protect\astroncite{{Van Horn} et~al.}{2018}]{vanhorn2018}
{Van Horn}, G., O.~M. Aodha, Y.~Song, Y.~Cui, C.~Sun, A.~Shepard, H.~Adam,
  P.~Perona, and S.~Belongie\leavevmode\nopagebreak\newline 2018.
\newblock The {iNaturalist} species classification and detection dataset.
\newblock {\em The IEEE Conference on Computer Vision and Pattern Recognition
  (CVPR)}.

\bibitem[\protect\astroncite{Wagner}{2020}]{wagner2020}
Wagner, D.~L.\leavevmode\nopagebreak\newline 2020.
\newblock Insect declines in the anthropocene.
\newblock {\em Annual Review of Entomolgy}, 65.

\bibitem[\protect\astroncite{Weeks et~al.}{1997}]{weeks1997}
Weeks, P. J.~D., I.~D. Gauld, K.~J. Gaston, and M.~A.
  {O'Neill}\leavevmode\nopagebreak\newline 1997.
\newblock Automating the identification of insects: a new solution to an old
  problem.
\newblock {\em Bulletin of Entomological Research}, 87(2):203--211.

\bibitem[\protect\astroncite{Zhang et~al.}{2010}]{zhang2010}
Zhang, X., Y.~Gao, and T.~Caelli\leavevmode\nopagebreak\newline 2010.
\newblock Primitive-based {3D} structure inference from a single {2D} image for
  insect modeling: towards an electronic field guide for insect identification.
\newblock {\em 11th International Conference on Control Automation Robotics \&
  Vision}, IEEE.

\bibitem[\protect\astroncite{Zimmermann et~al.}{2015}]{zimmermann2015}
Zimmermann, J., G.~Glockner, R.~Jahn, N.~Enke, and
  B.~Gemeinholzer\leavevmode\nopagebreak\newline 2015.
\newblock Meta-barcoding vs. morphological identification to assess diatom
  diversity in environmental studies.
\newblock {\em Molecular Ecology Resources}, 15:526--542.

\end{thebibliography}

\newpage

\section*{Appendix}

\begin{table}[htbp]
	\caption{Average classification accuracy (and standard deviation) of test specimens for nine camera setting combinations (exposure $=\left[1000\mu\text{s},1500\mu\text{s},2000\mu\text{s}\right]$ and aperture $=\left[1:3.8,1:8,1:16\right]$). The highest classification accuracy is marked in bold.}
	\begin{center}
		\begin{tabular}{|c|rr|rr|rr|rr|rr|rr|}
			\hline
			& \multicolumn{6}{c|}{Resnet-50} & \multicolumn{6}{c|}{InceptionV3}\\
			& \multicolumn{2}{c}{1000} & \multicolumn{2}{c}{1500} & \multicolumn{2}{c|}{2000} 	& \multicolumn{2}{c}{1000} & \multicolumn{2}{c}{1500} & \multicolumn{2}{c|}{2000}\\
			\hline
			1:3.8 & 0.952 & (0.025) & 0.956 & (0.021) & 0.955 & (0.022) & 0.956 & (0.020) & 0.951 & (0.023) & 0.945 & (0.026)\\
			1:8 & 0.949 & (0.025) & 0.954 & (0.017) & 0.960 & (0.012) & 0.945 & (0.024) & 0.954 & (0.016) & \textbf{0.964} & (0.015)\\
			1:16 & 0.940 & (0.019) & 0.955 & (0.017) & 0.945 & (0.024) & 0.940 & (0.015) & 0.944 & (0.025) & 0.945 & (0.024)\\
			\hline
		\end{tabular}
	\end{center}
	\label{weighted_sum}
\end{table}

\end{document}